\documentclass[runningheads]{llncs}
\usepackage{graphicx}
\usepackage{multicol,caption}
\usepackage{float}
\begin{document}
\title{Ship Detection: Parameter Server Variant}
%
%
\author{Benjamin Smith}

%
\authorrunning{B. SMITH}
%
\institute{University of Victoria
\email{benjaminsmith@uvic.ca}\\
}
%
\maketitle              
\begin{abstract}


Deep learning ship detection in satellite optical imagery suffers from false positive occurrences with clouds, landmasses, and man-made objects that interfere with correct classification of ships, typically limiting class accuracy scores to 88\%. 
This work explores the tensions between customization strategies, class accuracy rates, training times, and costs in cloud based solutions. 
We demonstrate how a custom U-Net can achieve 92\% class accuracy over a validation dataset and 68\% over a target dataset with 90\% confidence.  
We also compare a single node architecture with a parameter server variant whose workers act as a boosting mechanism. 
The parameter server variant outperforms class accuracy on the target dataset reaching 73\% class accuracy compared to the best single node approach.
A comparative investigation on the systematic performance of the single node and parameter server variant architectures is discussed with support from empirical findings. 

\end{abstract}
\section{Introduction}
    The ability to detect ships in optical satellite imagery provides issues surrounding clouds, landmasses, man-made objects, and highly reflective objects that introduce false positives. 
This has been noted as a significant area of interest within the remote sensing community \cite{Hordiiuk_2019} \cite{Ma_2019}. 
In this work, the goal is to exceed related works accuracy in the domain of ship detection using optical satellite imagery and investigate systematic benchmarks between two architectural approaches. 
Using a training/validation dataset distribution matched to that of a target distribution, a custom U-Net model will be trained in order to predict segmentation masks of ships of varying sizes. 
The target dataset is provided by industry partner UrtheCast, Vancouver, B.C. from their Theia MRC satellite. 
Since no ship detection dataset exists with the target dataset, a dataset taken by the SPOT-5 satellite will be used for training. 

This custom U-Net model implements deeper layers and bottlenecks that aim to provide a better feature representation during the encoding phase and result in a improved class accurate segmentation mask than other works. 
This claim will be supported empirically through class accuracy scores on validation and target data. 
The custom U-Net will be trained with two different systematic approaches: a single node architecture that closely relates to standard training and a parameter server variant. 
This variant implements the workers as weak learner boosters via gradient accumulation in order to train over the dataset. 
The parameter server variant aims to explore multiple gradient paths via multiple workers and result in a more generalizable model.
The outcome of increased class accuracy detecting ships in satellite optical imagery can provide beneficial use-case applications for industry and governmental bodies in areas of economic, ecological, and security domains.

The rest of the paper is structured as follows: Background and Related Works, Methodology, Experiments, Discussion, and finally Conclusion. 
\section{Background and Related Works}
    
\subsection{Ship Detection}
    
In regard to maritime object detection, the more generic term of "vessel" is unused as it refers to other objects that are not distinctly ships, such as floating docks and canoes.  
Locating ships in aerial imagery is typically performed using Synthetic Aperture Radar (SAR) data \cite{Karakus_2020} 
\cite{KANJIR20181}. 
SAR receives high degrees of accuracy since the data provided are not affected by sunlight or clouds. 
However, utilizing satellite optical imagery provides more identifiable information about ships \cite{KANJIR20181}. 
Ship detection in coastal and open bodies of water has a wide range of applications such as economic, security, environmental, and ecological \cite{KANJIR20181}. 
Previous work implementing object segmentation models (focused on U-Net) have seen accuracies reach ~57\% \cite{Siam_2018} and for the same dataset used in this work, ~85\% \cite{Hordiiuk_2019}.
Another approach implements other model types, as well as their own, that are typically used for object detection (YOLO, Faster R-CNN, SSD, etc) ranged from 62.4\% to 88.3\% accuracies \cite{Ma_2019}.

\subsection{Distributed Scheduling}
    Distributed systems can be utilized in order to improve the computational power of a system \cite{Yang_2018}. 
By horizontally scaling the number of compute nodes, the amount of computational power increases. 
This allows systems to process a larger amount of data/tasks as compared to a single machine. 
Vertically scaling is typically more expensive, thus the focus on horizontal scaling. 
With the increase of nodes and compute power, there is also an increase in the issues surrounding distributed systems.

The general scheduling problem can be written as $\alpha | \beta | \gamma | \mu$ \cite{Bittencourt_2018}.
The first attribute, $\alpha$, is a single entry and description of the execution environment. 
The attribute $\beta$ can have from zero to multiple entries, and describes what is the processing to be done. 
The $\gamma$ attribute details the description of the objective to be optimized.
Finally, $\mu$ is the charging criteria of models adopted in the system: hourly based charge, an auction system, per-transaction charge, data-access charge, and so on. 

By distributing computation, there is a need for oversight and regularization of tasks for the system to process, this is where scheduling comes in. 
Scheduling is a decision-making process that is used on a regular basis in many manufacturing and services industries. 
It deals with the allocation of resources to tasks over a given time period and its goal is to optimize one or more objectives \cite{polino2018model}.
Implementing a distributed system requires cost for hardware support and agreements on service expectations. 
Optimizing tasks through proper scheduling helps reduce the overall cost of computation while increasing the value customers receive.
\subsection{Architecture}
    
A system is centralized when it contains a central node or process that performs the necessary delegations to other nodes — this node is called the primary node \cite{Bittencourt_2018}. 
The other delegated nodes are considered workers. 
Decentralized distributed systems offer a more obscure abstraction of primary/worker relations. 
A node may be a primary of a centralized subset of the network and simultaneously exist as a worker of the larger network, or it may be a worker in its entirety while existing at the same level as a primary with its own worker pool. 
The primary nodes communicate between each other in order to achieve the system goal.

Another type of distributed system architecture falls into a hierarchical structure. 
This hierarchical distributed system offers a tiered layer where a node may communicate with only its predecessor or successor (either one level above or below).

The organization of the architecture may play a role in the scope of the scheduler. 
Not only should we consider the architecture connectivity of primary, workers, and their resources but also policies and access controls when implementing a scheduler. 
In a cloud based environment, these access controls limit the network and physical resources a worker can communicate with.
\subsection{Clusters}
    
A cluster, in this realm, is a collection of dedicated computational hardware that can communicate; usually over a dedicated private network (but distributed systems also exist over public networks with heightened security protocols) \cite{Bittencourt_2018}. 
Careful consideration towards task allocation can have immediate effects on the efficiency of a system. 
The type of system is what dictates the best scheduling approach. 
In a synchronous scheduling of tasks, the network is only as fast as its slowest worker. 
The same applies to asynchronous architectures, but typically with a faster rate of overall system completion. 
The rate of completion also depends on the nature of the task itself. 
If tasks are not dependent on each other, then the system may progress concurrently. 
However, if tasks do depend on each other, then the system may experience a varied progress rate where at times it may be zero.
\subsection{Tasks}
    
A task is a logical unit that makes progress toward a unified goal of a system \cite{bhatti_2017}. 
One such set of tasks could be images that a distributed model is learning or predicting on. 
Another such set of tasks could be a set of computations offloaded to multiple nodes in small chunks and sent off to workers to be processed. 
A task may end at the worker node — perhaps a state change — or a response may be communicated back to the primary.
There are two types of tasks: dependent and independent.

Dependent tasks rely on the response from a task ahead of itself before they can be executed. 
These tasks are better suited for a synchronous system that waterfalls from one task to the next depending on dependency. 
The system makes progress at a more determinable rate due to the predictive analysis of task succession.

Independent tasks can be run in parallel and/or concurrently. 
These tasks run best in an asynchronous system. 
The downfall to consider is resource contention if a particular set of tasks require the same network resource (say, accessing a file on a shared storage) — this situation has workarounds: eventual consistent databases/servers. 
The system makes unified progress at an undetermined rate as tasks may run simultaneously toward the system goal.
\subsection{Task Allocation Through The Scheduler}
    
The process of managing task allocation, to where and to whom, is the responsibility of a scheduler. 
The scheduler of a distributed system performs akin to the process scheduler on any operating system.

Task-graph scheduling is considered an NP-hard problem \cite{xie_shao_xin_2016}.
Scheduling tasks from the primary to workers can occur at multiple stages. 
There can be global scheduler that directs all tasks to all connected workers. 
There can also be local schedulers that handle incoming and outgoing tasks/replies on both the primary and worker nodes. 
Organizing this dynamic scale of events is a major component of the system efficiency. 
In a centralized distributed system, it is more common to see a global scheduler managing the allocation of resources and tasks between primary and workers. 
However, in a decentralized distributed system, there may exist multiple schedulers that are responsible for part of the overall system. 
Management of this decentralized state may be distributed over resources.
Each scheduler in a decentralized system can be considered the same as that of the scheduler in a centralized system.
Schedulers are considered to follow one of two main categories: List or Cluster scheduling \cite{bhatti_2017}. 
List scheduling is composed of two parts: a priority/ordering of tasks and a mapping to a processing node. 
Cluster scheduling extends List scheduling by mapping collections or sets to processing nodes. 

\subsection{Cloud-Based Scheduling}
    
Infrastructure as a Service (IaaS), Software as a Service (SaaS), and Platform as a Service (PaaS) are common models for user services and each have their own scheduling requirements \cite{Bittencourt_2018}.

IaaS — the most popular in cloud environments — offers physical and virtual environments to deploy computations and/or applications at scale on a public or private based setting. 
Some of these features are restricted through access controls that are subject to cost-per-use or subscription-based. 
For IaaS, the scheduler is responsible for “allocating virtual machine requests on the physical [architecture]” \cite{Bittencourt_2018}. 
For PaaS and SaaS, the scheduler is responsible for allocating the aforementioned applications and/or computational chunks to workers in order to be executed/served.

Part of optimizing the objectives of cloud-based scheduling includes the Service Level Agreement (SLA) of users with their end clients (workers).
The throughput and latency of a system directly affects users first and third party SLA agreements — these SLA agreements can be synonymous to Quality of Service (QoS) expectations.

To achieve high SLA/QoS, a scheduler needs to be robust in order to reduce execution time and cost. 
A robust scheduler must be able to handle a change in application topology (new instances dynamically being added and removed from the pool), resource/software configurations, input data, and data sources. 
This is not a comprehensive list of possible dynamics a scheduler should be robust to. 
One of the biggest nuances of distributed systems is the ability of a schedulers' robustness against failure occurrences.

Handling failure occurrences is a major proponent of distributed systems.
The scheduler must be able to salvage, redistribute, and/or re-implement these occurrences in order to maintain system QoS. 
Granted, if a failure occurrence is large enough, there will be interruptions in service. 
However, if a failure occurrence is small scale and handled properly, an end user may never even notice.
\subsection{Distributed Deep Learning Scheduling}
    
There are many kinds of scheduling algorithms that each satisfy different constraints of a system. 
Choosing the most effective algorithm ensures the optimization of the system goals are achieved.

During prediction, production models will most likely be served in an app on a scalable cluster. 
This cluster can be either on a heterogeneous or homogeneous system and tasks will be allocated with the above type algorithms.
However, Machine/Deep Learning task allocation during training are more specifically scheduled depending on the architecture of the framework. 
Framework architectures may be decentralized, such as AllReduce, or centralized, such as the Parameter Server. 
The following two algorithms are focused on the Parameter Server architecture. 
They aim to reduce the limitations imposed by straggling workers (workers that have fallen behind the others progress rate).
Such algorithms include Cyclic or Staircase Scheduling \cite{Amiri_2019}.
\subsection{Stragglers}
    
When provided with tasks, workers sometimes either become corrupt (maybe by a broken process) or another process has started (out of the control of the user and worker node) that demands more computational power that causes the computation of the task to take longer; these workers are referred to as \textit{stragglers}  \cite{Ozfatura_2019,Amiri_2019}. 
Perhaps it is the task itself that is computationally demanding that causes a worker to take longer. 
In other (probably most) cases, it is the network that bottlenecks task completion communication between the worker and primary \cite{Yang_2018}.

When one (or more) workers begin to experience these disruptions, they fall behind the other workers in terms of progress over assigned tasks — these are considered straggling workers. 
The product these straggling workers output may become stale if the application requires operations on current data and therefore useless. 
If stale data is used there is the possibility that it will hinder the progress made by other workers.

It’s fairly intuitive to reason about workers that are persistently straggling — they fell behind and are in a constant state of catch up and are persistently falling behind due to lack of computational power/network bottlenecking compared to other workers. 
Non-persistent types of straggling workers are harder to predict and analyze solutions for \cite{Ozfatura_2019,Amiri_2019}.
Non-persistent straggling workers are intermittently straggling but may not always be in as much of a deficit as compared to other workers as they may complete a significant portion of the assigned tasks by the time other workers have completed theirs.

\subsection{Parameter Server}
    
The architecture of a parameter-server system, typically centralized, can involve multiple layers of server groups, schedulers, and workers \cite{comm_eff_ps}. 
This architecture is used to leverage more compute power against massive datasets, such as datasets found in industry that can go from TBs to PTs \cite{comm_eff_ps}.
Attempts to provide better solutions for convergence with parameter servers involve a process that makes use of accumulated gradients \cite{accum_ps}. 
Parameter servers do not typically converge at the same rate as synchronous SGD approaches \cite{integradted_ps} \cite{Zhang_2020}. 
\section{Methodology}
    
Two architectures, single node and parameter server variant, were used to learn ship detection and tested on a target dataset. 

\label{training_ds}

\subsubsection{Training Dataset}

The term layers and channels are synonymous in regard to the following description about the datasets.  
A 48 thousand chip dataset was re-sasmpled to five-meter and quartered to a (3,384,384) chip (RGB) that each contained a ship and was taken by the SPOT-5 satellite \cite{spot-5}; originally built by AirBus \cite{airbus}.
Only 1.035e-6\% of the overall dataset contains ships which is a severe class imbalance. 
The remaining percentage of pixels are no-ship. 
To combat the imbalance, a Focal Dice Loss was implemented for the model which focuses more on the smaller classes within data for object segmentation \cite{focaldiceloss}. 

Originally the dataset contained 48 thousand ship chips and 128 thousand non-ship chips of shape (3,768,768). 
However, regarding ship and no-ship as the classes, the original dataset was severely imbalanced. 
As of this, the 128 thousand non-ship chips were discarded and the remaining 48 thousand were segmented into quarters - resulting in the 63 thousand chips of shape (3,384,384) that each contained ships. 
The chips were quartered and not resized as to retain the correct resolution.
Only 1.035e-6\% of the overall dataset contains ships which is a severe class imbalance. 
The remaining percentage of pixels are no-ship. 
To combat the imbalance, a Focal Dice Loss was implemented for the model which focuses more on the smaller classes within data for object segmentation \cite{focaldiceloss}. 
Additionally, the dataset underwent a random augmentation: vertical and horizontal flipping, or no augmentation during training. 

\subsubsection{Target Dataset}

A 79 chip target dataset is built using the optical imagery generated by Theia, the MRC satellite optical camera \cite{UC_theia} and is provided by industry partner UrtheCast, Vancouver, B.C. taken by their Theia MRC satellite over Burrard Inlet, Vancouver, BC, Canada.
It is a five meter resolution camera, meaning that each pixel represented five meters on the ground. 
This dataset consisted of 79 chips that were built from an MRC scene over Burrard Inlet, Vancouver, British Columbia.
Each datapoint in the target dataset was built into the shape (3,384,384). 
Theia MRC data has never been used for this use-case (ship detection).
This dataset was used for testing against an unseen satellite dataset in order to determine the generalizable capabilities of the trained model. 
Like the training/validation dataset, the ship class is severely imbalanced and only accounts for 0.2508\% of the dataset. 
\subsubsection{Custom U-Net}
    \label{unet}

\begin{figure}[!h]
    \centering
    \includegraphics[scale=0.35]{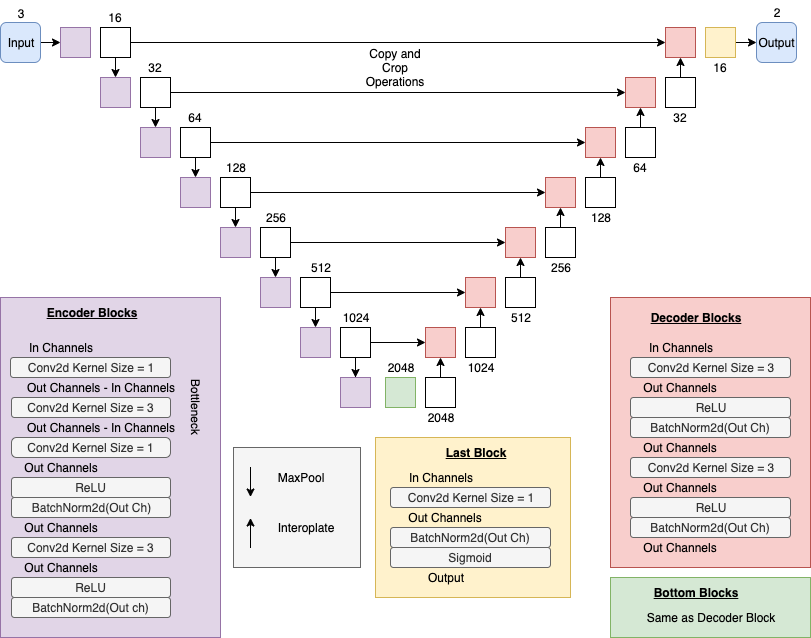}
    \caption{Custom U-Net Architecture}
    \label{fig:unet}
\end{figure}

The custom U-Net model in Figure \ref{fig:unet} was implemented with the PyTorch deep learning framework and can be thought about in two sections: encoder and decoder. 

The custom U-Net goes an extra layer deeper than the traditional U-Net model \cite{unet} designed to extract more features from the images. 
This increases the complexity of the model and aims to capture more distinct features in examples. The number of trainable parameters increase from 31,042,434 to 306,270,726, about 9.87 times more \ref{tab:unet_params}.
Also implemented are bottlenecks in the encoding blocks.
These bottlenecks are used to gather the best features of the distribution and therefore minimize loss. 

Regarding the encoder block bottlenecks, the input channels are subtracted from the output channels as a means of dimensionality compression to gain the most relevant features through a 1x1 kernel (or filter), then processed via a 3x3 kernel to the same number of reduced channels before being processed to the target number out channels via 1x1 kernel. 
The reduction in dimensionality before the expensive 3x3 kernel sized convolution helps reduce the computational resources (number of parameters to process) from increasing too much.

The numbers above each box in Figure \ref{fig:unet} indicate the number of channels/layers the tensor has at that layer of the U-Net model. 
As such, the input images are RGB and therefore 3 channels : (3,X,Y). 
This is run through an encoder block and transformed into a deeper feature tensor (3 channels to 16 from input to the first layer of the U-Net). 
This transformed layer is then put through a 'MaxPool' convolution to the next U-Net layer and the process repeats until the bottom of the U-Net where the features of the images are encoded into the 2048 channel tensor. 

The bottom layer at the end of the encoder holds the reduced dimensionality feature space of the input batch. 
It runs the tensor through a 3x3 kernel 2D convolutional layer twice to extract as many features as possible before being interpolated up a U-Net layer to the start of the decoding block. 
At each interpolation, the equally deep encoded layer is cropped and copied over (to enhance the segmentation to the objects that are to be detected) and then put through the decoder block convolutions. 
This process is repeated up until the top layer of the U-Net model where the tensor is refined through a 1x1 kernel 2D convolutional layer to the number of classes requested and passed through the sigmoid function to normalize the output values between 0 and 1.
This gives a probability for each pixel of what the U-Net thinks is the desired object class.
The architecture used in this research utilized stochastic gradient descent as the optimization method for model convergence.

\begin{table}[!h]
    \centering
    \caption{Trainable Parameters of U-Net Versions}
    \begin{tabular}{|c|c|}
        \hline
        Model & Trainable Parameters \\
        \hline
        Original U-Net & 31,042,434 \\
        \hline
        Custom U-Net & 306,270,726 \\
        \hline
    \end{tabular}
    \label{tab:unet_params}
\end{table}
\subsection{RedisAI as a Data Server}
    Delivering data to the training cycle was handled by a Redis connection client. 
A module of Redis, RedisAI \cite{redisai}, that within it's datatype structure can manage tensors via the Redis standard key-value approach is used as the server structure. 
This is useful to communicate large amounts of tensors back and forth asynchronously since Redis can handle massive amounts of requests - it can easily handle 10,000 clients as of Redis version 2.6 \cite{redis}. 
Leveraging RedisAIs in-memory caching system allows for training data to be delivered rapidly. 
One downfall to this approach is that task delivery is limited by the network bandwidth as with all network bound communications. 
However, this delivery service allowed for the training data to be available and in a ready format in memory without the use of file I/O operations. 
was particularly useful as the single node architecture was trained concurrently with the following parameter server variant with both using the same data server. 
Also, this method of data delivery allowed for concurrent reads of key-values. 
This means that, in a parameter server architecture, any worker may read the global model weights concurrently as another worker.
This eliminates downtime due to resource locks. 
In an Azure virtual private network, specific inbound port rules have to be assured to prevent malicious interaction/use from outside users. 
\subsubsection{Single-Node Architecture}
    The single node architecture is a typical training scheme deployed over a single Azure NC24 virtual machine.
This virtual machine is host to four Tesla K80 GPU's with 12GB memory for each GPU (48 GB total) and 220 GB of RAM \cite{nc24}. 
To take advantage of four GPU's, the PyTorch data parallel functionality was implemented. 
This distributes the model across all four GPU's and delivers equally divided batches to each GPU. 
At the end of each epoch of training, the model was saved to a mounted Azure Fileshare. 
With this process, there is the upfront overhead of loading model replicas to each GPU, but once loaded they stay in GPU memory for the training session.
Retaining GPU memory allows for quick updates to GPU-local models. 
The GPU-local worker processes the batch and returns its gradients for which the main loop will average across all GPU's for the backward propagation of the model. 
This is a fairly standard approach to training deep learning models.
A RedisAI \cite{redisai} data server hosted all training data for this architecture. 
\subsubsection{Parameter Server Variant Architecture}
    
PSv architecture was compromised of five Microsoft Azure NC series virtual machines connected over a private virtual network. 
Each one mounting the same Fileshare for model and metric save points.
It was implemented as a centralized parameter server, where one primary node hosted the global model which the registered workers would implement gradient sharing for. 
At the end of each epoch of training, the model was saved to a mounted Azure Fileshare. 
A RedisAI \cite{redisai} data server hosted all training data as well as global model weights and worker model gradients for this architecture. 

The primary node was implemented on a NC24 virtual machine which hosted four Tesla K80 GPU's, just like the single node architecture. 
This VM also hosted the the scheduler and the data server (due to the large amount of RAM available it was cheaper than deploying on a memory optimized virtual machine or separate VMs for each process) as well as the primary process.
Each worker was implemented on a NC6 virtual machine which hosted a single Tesla K80 GPU with 12 GB of GPU memory. 

Workers were leveraged as weak learner boosters that accumulated gradients to be passed to the primary process.
Once the primary process received gradients from a worker, it accumulated them with another forward pass mapping the gradients into a backward propagation function before global weight updates were performed by the optimizer. 
All weights and gradients were published and subscribed to on the data server and keyed respectively to the publishing worker. 
The primary process reads each workers gradients synchronously and updates the global model weights before publishing them back to the data server. 
The scheduler handled the tasks to workers. 
Each task contained a set of image keys that the worker could retrieve from the data server in order to build an input batch for the worker-local model. 
Communication was handled through Azure Tables. 
The scheduler also retained the progress made through the dataset. 
If a worker were to join later either by restart or as a new entity, that worker would simply receive the next appropriate set of keys.
The keys were shuffled at the end of each dataset iteration. 
\section{Experiments}
    
Three experiments were performed with the SN architecture and one with the PSv architecture. 
Class accuracies and systematic metrics will be recorded. 
\vspace{-4mm}

\subsection{Data Serving}
    
Both frameworks utilized a RediaAI server to load chips. 
One of the most effective ways of serving traditional files for deep learning is using a HDF5 format and served as a comparison to RedisAI. 
For this experiment, each method of delivery iterated through individual files (both identical Numpy array format) and converted into float tensors that a model could receive. 
However, the model step was skipped as to just compare file serving methods. 
The results are summarized in Table \ref{tab:file_serve}.

\begin{table}[!h]
    \small
    \centering
    \begin{tabular}{|c|c|}
        \hline
        Method & Time \\
        \hline
        RediaAI & 0:02:32.101122 \\
        \hline
        HDF5 I/O & 3:37:02.746008 \\
        \hline
    \end{tabular}
    \caption{File Serving Time Comparison}
    \label{tab:file_serve}
\end{table}
It is intuitive that utilizing the data server is the more effective approach since it processed all 63 thousand chips within a few minutes, whereas HDF5 I/O took over three hours. 
The use of a data server increases the usage of the GPU to 95-99\% consistently as per the nvidia-smi command, whereas file I/O would see dips to 0\% (although microseconds) when workers are not able to process an item or are limited by the number of cores versus the batch size in relation to the overhead of file I/O. 
\subsubsection{Single-Node Experiments}
    
During training, hyperparameters for Focal Dice Loss and a learning rate scheduler were adjusted through micro-testing. 
The Focal Dice Loss hyper parameters alpha and beta were set to 10.0 and 2.0 respectively. 
When the alpha parameter was lower than 10.0, the loss function would not pick up on the smaller class size. 
The learning rate was set to 1e-2, momentum to 0.9, and weight decay to 1e-5 as the hyperparameters for training.
A PyTorch cyclic learning rate scheduler was also implemented with the minimum as 1e-2 and the maximum as 1e-1 in order to prevent the model from getting stuck in local minimums. 
The SN architecture was trained over dataset sizes of 10 and 20 thousand for 30 epochs, and there was also a run with dataset size of 10 thousand over 100 epochs.
The focus on using a subset of the data is that of time and cost constraints using virtual machines in the Azure cloud space.
In the following tables, each training approach will be keyed as such: \textbf{A}: 10k dataset and 30 epochs, \textbf{B}: 20k dataset and 30 epochs, and \textbf{C}: 10k dataset and 100 epochs. 
The results of training are summarized in Table \ref{tab:lf_loss_iou}.
Over the training cycle, various metrics were recorded to give insight to the performance of the architecture. 
One metric involved the time in seconds it took to complete each batch, seen in Table \ref{tab:batch_train}.
This process involves retrieving the batch items, forward passing through the U-Net model, calculating loss and accuracy, performing a backward pass through the U-Net model to calculate its gradients, and finally performing an optimization step to update the U-Net weights ready for the next batch.

\vspace{-4mm}
\begin{multicols}{2}
\begin{table}[H]
    \small
    \centering
    \caption{SN Loss and IoU Scores}
    \begin{tabular}{|c|c|c|}
         \hline
         Approach & Loss & IoU  \\
         \hline
         A & 5.83 & 47.8\% \\
         \hline
         B & 5.83 & 52.8\% \\
         \hline
         C & 5.80 & 56.2\% \\
         \hline
    \end{tabular}
    \label{tab:lf_loss_iou}
    \vspace{-4mm}
\end{table}
\columnbreak
\begin{table}[H]
    \small
    \centering
    \caption{Time Taken to Prepare and Train a Single Batch (seconds)}
    \begin{tabular}{|c|c|c|c|c|}
         \hline
         Approach & Min & Mean & Max & Variance \\
         \hline
         A & 2.06 & 2.16 & 83.19 & 0.82 \\
         \hline
         B & 2.07 & 2.16 & 84.83 & 0.45 \\
         \hline
         C & 2.08 & 2.18 & 84.93 & 0.35 \\
         \hline
    \end{tabular}
    \label{tab:batch_train}
    \vspace{-4mm}
\end{table}
\end{multicols}

Table \ref{tab:epoch_train} shows the metrics across approaches for how long in seconds it takes to iterate through the entire dataset once. 
At each epoch end, the dataset is shuffled so the model avoids overfitting. 

\begin{table}[!h]
    \small
    \centering
    \caption{Time Taken to Train a Single Epoch (seconds)}
    \begin{tabular}{|c|c|c|c|c|}
         \hline
         Approach & Min & Mean & Max & Variance \\
         \hline
         A & 784.84 & 676.23 & 670.71 & 189.01 \\
         \hline
         B & 1342.67 & 1351.55 & 1426.31 & 203.61 \\
         \hline
         C & 675.92 & 682.85 & 756.77 & 61.51 \\
         \hline
    \end{tabular}
    \label{tab:epoch_train}
    \vspace{-4mm}
\end{table}

At the end of each epoch, the model is saved as a measure of redundancy if anything were to go wrong. 
Table \ref{tab:save_model} shows how long it takes in seconds to save the model to the mounted Azure Fileshare via 'torch.save'.
To accompany IoU accuracies, Table \ref{tab:SN_conf_m} show the confusion matrices class accuracy for inference. 

\begin{multicols}{2}
\begin{table}[H]
    \small
    \centering
    \caption{Time Taken to Save Model (seconds)}
    \begin{tabular}{|c|c|c|c|c|}
         \hline
         Approach & Min & Mean & Max & Variance \\
         \hline
         A & 38.31 & 56.46 & 90.13 & 188.04 \\
         \hline
         B & 31.43 & 59.93 & 111.16 & 405.93 \\
         \hline
         C & 31.90 & 52.73 & 104.51 & 174.51 \\
         \hline
    \end{tabular}
    \label{tab:save_model}
    \vspace{-4mm}
\end{table}
\columnbreak
\begin{table}[H]
    \small
    \centering
    \caption{Model Training Run Confusion Matrices}
    \begin{tabular}{|c|c|c|c|c|c|}
         \hline
         Approach & TP & TN & FP & FN & Accuracy  \\
         \hline
         A & 79\% & 100\% & 21\% & 0.49\% & 89.27\% \\
         \hline
         B & 81\% & 100\% & 19\% & 0.45\% & 90.41\% \\
         \hline
         C & 81\% & 100\% & 16\% & 0.25\% & 91.87\% \\
         \hline
    \end{tabular}
    \label{tab:SN_conf_m}
    \vspace{-4mm}
\end{table}
\end{multicols}
\subsubsection{Parameter Server Variant Experiment}
    
The PSv was run over 30 epochs using the dataset size of 10 thousand points and implemented gradient accumulation four times at each of the four workers before they pushed gradients. 
The reason only four batch gradient accumulations were chosen was because 8 and 16, although allowed the model to iterate through batches much faster, resulted in poorer convergence. 
Only 10 thousand datapoints were used (identical to that used in the SN architecture) as to reduce time and cost of the training run. 
Due to the boosting nature of this architecture, a learning rate of 1e-3 was implemented, as well as momentum of 0.9 and weight decay of 1e-5 like the SN architecture. 
The same hyperparameters were also used for the Focal Dice Loss as the SN architecture. Table \ref{tab:psv_loss_iou} shows the loss and IoU accuracy for the training run. 
The IoU accuracy, gathered over model evaluation, achieved ~14\%, however still showed a slow decrease in loss value and thus still converging. 
The workers reach a training loss of around 1.28 and a validation loss around 6.54. 
The confusion matrix seen in Table \ref{tab:psv_conf_m} details an accuracy of 85.96\% across classes. 

\vspace{-4mm}
\begin{multicols}{2}
\begin{table}[H]
    \small
    \centering
    \caption{PSv Loss and IoU Scores}
    \begin{tabular}{|c|c|c|}
    \hline
         Entity & Loss & IoU \\
         \hline
         Training & 1.28 & N/A \\
         \hline
         Validation & 6.54 & 14\% \\
         \hline
    \end{tabular}
    \label{tab:psv_loss_iou}
    \vspace{-4mm}
\end{table}
\columnbreak
\begin{table}[H]
    \small
    \centering
    \caption{PSv Confusion Matrix}
    \begin{tabular}{|c|c|c|c|c|c|}
         \hline
         Training Run & TP & TN & FP & FN & Accuracy  \\
         \hline
         PSv & 72\% & 99\% & 28\% & 0.51\% & 85.96\% \\
         \hline
    \end{tabular}
    \label{tab:psv_conf_m}
    \vspace{-4mm}
\end{table}
\end{multicols}

At the beginning and after each consecutive gradient push, the worker will wait and then read the global models updated weights in order for the next set of batches to be processed. 
Table \ref{tab:read_global_model} shows the time taken in seconds for that initial and consecutive reads. 
Before a worker is able to process a batch, it must first request the keys from the scheduler from the current point of the dataset. 
Table \ref{tab:request_task} shows the time it takes in seconds for the worker to put in a request for a new batch to the Azure table the scheduler reads. 

\vspace{-4mm}
\begin{multicols}{2}
\begin{table}[H]
    \small
    \centering
    \caption{Time Taken to read Global Model (seconds)}
    \begin{tabular}{|c|c|c|c|c|}
        \hline
        Entity & Min & Mean & Max & Variance  \\
        \hline
        Worker\_001 & 18.58 & 22.36 & 30.65 & 3.25 \\
        \hline
        Worker\_002 & 18.35 & 22.49 & 31.30 & 2.83 \\
        \hline
        Worker\_003 & 19.45 & 22.70 & 31.46 & 4.41 \\
        \hline
        Worker\_004 & 18.48 & 21.97 & 31.24 & 2.89 \\
        \hline
    \end{tabular}
    \label{tab:read_global_model}
    \vspace{-4mm}
\end{table}
\columnbreak
\begin{table}[H]
    \small
    \centering
    \caption{Time Taken to Request Task (seconds)}
    \begin{tabular}{|c|c|c|c|c|}
        \hline
        Entity & Min & Mean & Max & Variance  \\
        \hline
        Worker\_001 & 0.035 & 0.101 & 1.35 & 0.008 \\
        \hline
        Worker\_002 & 0.035 & 0.099 & 1.23 & 0.008 \\
        \hline
        Worker\_003 & 0.034 & 0.098 & 3.80 & 0.009 \\
        \hline
        Worker\_004 & 0.036 & 0.094 & 1.99 & 0.007 \\
        \hline
    \end{tabular}
    \label{tab:request_task}
    \vspace{-4mm}
\end{table}
\end{multicols}

The scheduler may be busy processing another workers batch, so the requesting worker will have to wait until theirs in processed. 
Table \ref{tab:wait_for_task} shows the time in seconds each worker is waiting for their batch request to be processed by the scheduler. While the workers wait for their requests to be processed, the scheduler is busy performing other requests.
Table \ref{tab:deliverTask} shows the time in seconds taken by the scheduler to receive, process, and deliver the task back to a requesting worker. 

\vspace{-4mm}
\begin{multicols}{2}
\begin{table}[H]
    \small
    \centering
    \caption{Time Taken to Wait for Task (seconds)}
    \begin{tabular}{|c|c|c|c|c|}
        \hline
        Entity & Min & Mean & Max & Variance  \\
        \hline
        Worker\_001 & 0.012 & 0.080 & 1.24 & 0.006 \\
        \hline
        Worker\_002 & 0.014 & 0.078 & 0.90 & 0.006 \\
        \hline
        Worker\_003 & 0.012 & 0.079 & 3.78 & 0.008 \\
        \hline
        Worker\_004 & 0.014 & 0.073 & 0.74 & 0.005 \\
        \hline
    \end{tabular}
    \label{tab:wait_for_task}
    \vspace{-4mm}
\end{table}
\columnbreak
\begin{table}[H]
    \small
    \centering
    \caption{Time Taken for Scheduler to Process Request (seconds)}
    \begin{tabular}{|c|c|c|c|c|}
         \hline
         Entity & Min & Mean & Max & Variance \\
         \hline
         Scheduler & 0.00025 & 0.0135 & 3.74 & 0.0024 \\
         \hline
    \end{tabular}
    \label{tab:deliverTask}
    \vspace{-4mm}
\end{table}
\end{multicols}

Once the worker has received the task, it must prepare it for model input. 
Table \ref{tab:build_input} shows the time in seconds it takes for that preparation and model processing of that batch.
Once processed and accumulated over the four batches, the worker will push the local models gradients to the data server for the primary process to read. 
Table \ref{tab:update_model} shows the time in seconds it takes to push those gradients to the data server. 

\vspace{-4mm}
\begin{multicols}{2}
\begin{table}[H]
    \small
    \centering
    \caption{Time Taken to Build and Process Input (seconds)}
    \begin{tabular}{|c|c|c|c|c|}
        \hline
        Entity & Min & Mean & Max & Variance  \\
        \hline
        Worker\_001 & 0.498 & 1.16 & 2.53 & 0.068 \\
        \hline
        Worker\_002 & 0.116 & 1.11 & 3.05 & 0.064 \\
        \hline
        Worker\_003 & 0.600 & 1.21 & 2.76 & 0.087 \\
        \hline
        Worker\_004 & 0.333 & 1.12 & 2.19 & 0.065 \\
        \hline
    \end{tabular}
    \label{tab:build_input}
    \vspace{-4mm}
\end{table}
\columnbreak
\begin{table}[H]
    \small
    \centering
    \caption{Time Taken to Push Gradients From Worker (seconds)}
    \begin{tabular}{|c|c|c|c|c|}
        \hline
        Entity & Min & Mean & Max & Variance  \\
        \hline
        Worker\_001 & 5.37 & 9.27 & 35.40 & 26.23 \\
        \hline
        Worker\_002 & 5.35 & 9.00 & 34.41 & 21.38 \\
        \hline
        Worker\_003 & 5.36 & 9.68 & 35.72 & 32.40 \\
        \hline
        Worker\_004 & 5.21 & 8.99 & 30.81 & 23.93 \\
        \hline
    \end{tabular}
    \label{tab:update_model}
    \vspace{-4mm}
\end{table}
\end{multicols}

Once pushed, the worker will wait until the primary process has read, updated the global model, and pushed the updated weights to the data server ready for the worker to read and use in the next cycle. 
Table \ref{tab:wait_for_model_updates} shows the time in seconds the worker waits for the primary to execute that process. 
While the worker is waiting, the primary process reads and updates the global model.
Table \ref{tab:update_global_model} shows the time in seconds it takes for the primary process to perform that action.

\vspace{-4mm}
\begin{multicols}{2}
\begin{table}[H]
    \small
    \centering
    \caption{Time Taken to Wait for Global Weight Updates (seconds)}
    \begin{tabular}{|c|c|c|c|c|}
        \hline
        Entity & Min & Mean & Max & Variance  \\
        \hline
        Worker\_001 & 12.64 & 19.34 & 54.37 & 30.62 \\
        \hline
        Worker\_002 & 12.41 & 19.89 & 52.98 & 27.71 \\
        \hline
        Worker\_003 & 12.29 & 18.35 & 56.24 & 34.82 \\
        \hline
        Worker\_004 & 12.23 & 19.13 & 66.68 & 41.70 \\
        \hline
    \end{tabular}
    \label{tab:wait_for_model_updates}
    \vspace{-4mm}
\end{table}
\columnbreak
\begin{table}[H]
    \small
    \centering
    \caption{Time Taken to Read Gradients and Update Global Model (seconds)}
    \begin{tabular}{|c|c|c|c|c|}
        \hline
        Entity & Min & Mean & Max & Variance \\
        \hline
        Primary & 5.41 & 8.82 & 14.75 & 0.367 \\
        \hline
    \end{tabular}
    \label{tab:update_global_model}
    \vspace{-4mm}
\end{table}
\end{multicols}

Once updated, the global model updated weights are then pushed to the data server and the worker is then signalled to read those weights. 
Table \ref{tab:push_global_weights} shows the time in seconds it takes to push those updated weights and signal the worker. 
This process encapsulates the sub-processes of an epoch. 
Table \ref{tab:psv_epoch} shows the time records in seconds for the time taken to complete epochs over the entire 30 epoch training run. 
\vspace{-4mm}
\begin{table}[!h]
    \small
    \centering
    \caption{Time Taken Per Epoch Completion (seconds)}
    \begin{tabular}{|c|c|c|c|c|}
        \hline
        Entity & Min & Mean & Max & Variance  \\
        \hline
        Worker\_001 & 2432.45 & 2660.84 & 3064.24 & 39696.18 \\
        \hline
        Worker\_002 & 2425.73 & 2672.78 & 3040.29 & 42061.82 \\
        \hline
        Worker\_003 & 2441.96 & 2677.38 & 3213.13 & 50324.72 \\
        \hline
        Worker\_004 & 2413.94 & 2628.33 & 3016.29 & 27690.05 \\
        \hline
    \end{tabular}
    \label{tab:psv_epoch}
    \vspace{-4mm}
\end{table}

At the end of each epoch the scheduler will initialize a save for each worker to allow for redundancy if the primary were to fail. 
Table \ref{tab:saving_model:torch} shows the time in seconds it takes to save the local model with the workers most recent weight reads to disk. 

\vspace{-4mm}
\begin{multicols}{2}
\begin{table}[H]
    \small
    \centering
    \caption{Time Taken to Push Updated Global Weights to Data Server (seconds)}
    \begin{tabular}{|c|c|c|c|c|}
         \hline
         Entity & Min & Mean & Max & Variance \\
         \hline
         Primary & 2.98 & 4.76 & 7.55 & 0.97 \\
         \hline
    \end{tabular}
    \label{tab:push_global_weights}
    \vspace{-4mm}
\end{table}
\columnbreak
\begin{table}[H]
    \small
    \centering
    \caption{Time Taken to Save Local Model via torch.save (seconds)}
    \begin{tabular}{|c|c|c|c|c|}
        \hline
        Entity & Min & Mean & Max & Variance  \\
        \hline
        Worker\_001 & 15.30 & 25.25 & 42.52 & 56.81 \\
        \hline
        Worker\_002 & 13.35 & 30.49 & 56.78 & 142.78 \\
        \hline
        Worker\_003 & 12.34 & 25.88 & 46.31 & 59.27 \\
        \hline
        Worker\_004 & 16.48 & 32.48 & 54.51 & 82.29 \\
        \hline
    \end{tabular}
    \label{tab:saving_model:torch}
    \vspace{-4mm}
\end{table}
\end{multicols}

\subsection{System Completion}
    Table \ref{tab:sys_time} states the completion time in seconds and cost for all run experiments. 
The SN architecture made use of a single NC24 virtual machine for training. 
The PSv made use of a NC24 and four NC6 virtual machines. 
However, the data server was hosted on the primary node of the PSv  architecture - this was a cheaper solution than having it run on a individual virtual machine. 
Since the SN architecture used the data server, it can also be suggested that the virtual machine that it trained on could also host the data server in an isolated training cycle and would absorb the cost. 
The cost is calculated based on hourly usage of each virtual machine \cite{vm_price}. 
In these experiments, the promotional version of the virtual machines were used for a reduced cost while they were available, but the cost will also be calculated for full-price usage. 
The NC6 Promo cost \$0.396/hr and NC24 Promo \$1.584/hr. 
The non-Promo NC6 cost \$0.90/hr and non-Promo NC24 cost \$3.60/hr. 

\vspace{-4mm}
\begin{table}[!h]
    \small
    \centering
    \caption{Training System Completion Time and Cost}
    \begin{tabular}{|c|c|c|c|}
         \hline
         Approach & System Time & Promo Cost & Full Cost  \\
         \hline
         SN (10k/30ep) & 06:24:47.519815 & \$10.16 & \$23.09 \\
         \hline
         SN (20k/30ep) & 12:04:27.297251 & \$19.13 & \$43.47 \\
         \hline
         SN (10k/100ep) & 21:26:59.642056 & \$33.98 & \$77.22 \\
         \hline
         PSv (10k/30ep/4accum) &  22:10:41.253272 & \$70.26 & \$156.68 \\
         \hline
    \end{tabular}
    \label{tab:sys_time}
    \vspace{-4mm}
\end{table}
\subsubsection{Theia MRC Predictions}
    The 79 chip (3,384,384) was tested with each of the four architectures: SN, PSv, and their accompanying OffHEM versions. 
Table \ref{tab:mrc_results} shows the percentages of the true positives, true negatives, false positives, and false negative of the prediction versus the ground truth. 
In the final column is the overall accuracy. 
These columns are calculated using a confusion matrix. 

\vspace{-4mm}
\begin{table}[!h]
    \centering
    \caption{Theia MRC Test Results}
    \begin{tabular}{|c|c|c|c|c|c|c|}
         \hline
         Architecture & TP & TN & FP & FN & Accuracy & IoU mAP \\
         \hline
         SN & 36\% & 100\% & 64\% & 0.26\% & 68.01\% & 15.03\% \\ 
         \hline
         PSv & 47\% & 100\% & 53\% & 0.38\% & 73.37\% & 19.15\% \\
         \hline
    \end{tabular}
    \label{tab:mrc_results}
    \vspace{-4mm}
\end{table}
    
Three experiments with the SN architecture produced a 92\% accuracy and one experiment with the PSv reaching 86\% ship detection accuracy. 
Though lower accuracy, the accuracy scoring was opposite on the target dataset with SN being lower than the PSv. 
A multitude of systematic metrics were recorded during the training process. 
\section{Discussion}
    \subsection{Data Serving}
    
Since I/O operations are vastly more expensive than in-memory retrieval of data, this is why a data server was chosen over standard image reads for serving up datapoints. 
Iterating and transforming over the entire 63223 image chips and converting them into learning-ready tensors only took 2 minutes and 32 seconds while the file I/O operations took over 3 hours. 
This comparison to read and prepare is the reason serving data to models from a data server was the chosen approach. 
Also, utilizing a data server allowed for uninterrupted GPU usage that allows for a more cost effective approach to training. 
\subsection{Single-Node}

It is expected that each of the SN approaches achieve similar times for their functionalities. 
Consideration toward the batch training time needs to be taken in regard to the maximum and the variance. 
The small variance indicates that the average batch training time is consistently around the mean time. 
However, the maximum time is excessively larger than the mean. 
This is due to the preparation of the dataset at the beginning of each epoch as the architecture implements the built-in (PyTorch) data loader objects, which results in a larger up front time overhead. 
Intuitively, training on a 20k dataset takes around twice as long as when training with a 10k dataset. 
The timing records for the 10k dataset over 100 epochs are lower than the other two most likely because the total number of recordings are numerously greater than the 30 epoch training runs. 
Though, it is interesting that the variance of the time it took the 20k dataset over 30 epoch training run to save the model to the Azure Fileshare is quite large compared to the other two approaches. 
The high variance of model saving could suggest that there was more fluctuation in network traffic between the compute node and the storage location or internal process contention at the time of this training run and at the end of each epoch when the model is saved since the standard deviation is about 11 seconds for Worker\_002 and about 7-9 seconds for the other workers. 
Though, it is more likely there was internal process contention delaying the initialization of the saving process within the VM that may also be paired alongside network bandwidth that causes such a large discrepancy. 
That same hypothesis could be applied to the other two approaches also at a much lesser degree since the variance is not very small. 
\subsection{Parameter Server Variant}
    
Reading of the global model to initialize the local models is at the beginning of the PSv pipeline.
The variance of records is small enough to indicate that there were no major networking or local issues at each read. 
Reading of the global model is one of the larger portions of overhead in the system. 
However, operating with a Redis based data server allows for concurrent reads and therefore reducing contention around resources.
This larger time is indicative to the models large size for this application. 

When the scheduler receives a task request, it delivers the task to a worker which processes it into a batch before performing a forward pass through the model. 
While the variance across all workers is really small indicating that the mean value is close to the true time, it is notable that worker\_003 maximum value was recorded as three times as long as the other workers. 
This is most likely a rare event (considering the variance is so small), and probably due to some internal process contention. 
The process of waiting for the task to be delivered and the time taken to process a task request are interoperable and dependent. 
A reason the worker wait times are bigger than the scheduler process times are due to the scheduler processing worker tasks synchronously as they are received and some workers are forced to wait while other workers are served. 
Once the worker has received the task, it is read from the data server, built into a batch, and a forward pass is processed. 
The times taken for processing a task, all with relatively small variances, are all faster than that of the SN architecture. 
Implementing a PyTorch data loader is convenient, however, it introduces an overhead. 
This step can be multiplied by the number of accumulations to represent a more accurate "batch" training time, but it is assessed individually with respect to the SN architecture performance comparison. 
Implementing a request-deliver type scheduling algorithm reduces contention over resources and the chances of stragglers occurring. 
Although it takes half as much time to prepare and process each batch, the accompanying overhead from requesting tasks and pushing/pulling global model updates causes the PSv architecture to operate slower than the SN architecture.

Uploading gradients for the large U-Net model varied over the training run as per the larger variance value in Table \ref{tab:update_model}. 
This deviation from the mean could be attributed to the complexity or size of the update or it could be network bound from uploading a large amount of data. 
Most likely, it is a combination of both of these things that cause the overhead. 
The minimum time for each worker suggests an ideal upload traffic load (close to zero) between the worker node and the primary node. 
Whereas the maximum time is the opposite and worst case scenario. 
Once pushed, the worker will begin waiting for the primary process to retrieve and update the global model weights before pushing them for the worker to read. 
The process is synchronous at the primary process as gradients are applied as they come in and recognized at each query. 
This means, that if a worker were to update just after the primary process queried the update table, then it would have to wait longer until the next query which would occur after the current query updates had been applied. 

This is reflective in the maximum wait times of the workers. 
The minimum wait time occurs when the primary process recognizes and implements that workers update first. 
Since this operation is synchronous and intermittent between worker update queries, this causes the variance to grow larger. 
The worst case scenario would be a worker having to wait for up to six updates before having theirs processed due to missing a query involving the other three workers and then having their update recognized at the bottom of the following query. 

A portion of this waiting is actually the primary process retrieving gradients, updating and pushing the global model weights. 
This is observed in the small variance between timing records for reading gradients and pushing weights from Tables \ref{tab:update_global_model} and \ref{tab:push_global_weights}.
These two functionalities are synchronous and the combined timing effects the workers wait time heavily. 

Each epoch takes around twice as long as the SN architecture, which is seen in Table \ref{tab:psv_epoch}, and varies significantly over the 30 epochs as per the extremely large variance value. 
All the timing effects discussed previously accumulate into this epoch completion time. 
There are multiple reads of the global model by workers, pushing gradients, and pushing weights that all accumulate into a larger overhead than the SN architecture. 
This architecture was performed using four workers with capabilities to scale beyond that. 
However, there would be a limit as the primary process would become more of a bottleneck than it currently is since workers must wait for their global model updates to complete. 
Again, like the SN architecture, at the end of each epoch each worker saves their recently pulled model to disk as a measure of redundancy in case there is a system failure. 
The reason that the PSv architecture takes half as long to save the respective model is the missing optimizing state dictionary that the SN architecture retains. 

Future work would be best served investigating efficient mechanisms to asynchronously have workers push gradients, primary update weights, primary push weights, and workers read weights on a layer-by-layer basis.
Eliminating the bottleneck of a centralized primary node may force considerations towards ring-AllReduce approaches for gradient sharing. 
Other approaches for efficiency increase could lie at the gradient accumulation stage. 
Determining the effects and overcoming these issues at larger accumulation rounds could increase the performance of epoch completion. 
However, it is unknown how these larger accumulation rounds would affect convergence given the current experiments.
\subsection{Cost}
    
The cost of running deep learning training for the SN architecture is relatively cheap. 
Especially for early stopped training runs or runs that do not require such complex data. 
The PSv approach cost over twice as much as the SN architecture within the same amount of time of training. 
This is obvious since the PSv architecture used more resources. 
Even using 5 virtual machines to run a distributed parameter server (variant), it is an affordable approach for distributed deep learning training given that the application justifies monetary expense. 
\subsection{Future Work}
    
Communication bottlenecks are a major concern for the PSv architecture. 
Overcoming the primary process bottleneck would require a tiered server architecture to overcome and therefore introducing complexity to the system. 
This involves having workers subscribe to a gradient accumulation server, which would then push to a higher level and pull down the available global model weights. 
This architecture may introduce a higher degree of weight staleness but could be considered a systematically efficient improvement over the current PSv architecture. 
Investigations into the effect of stale weights and object detection could be considered for future work.
Reducing systematic overhead within the PSv architecture would improve its efficiency in order to exceed standard training times. 
Within this same vein, implementing an asynchronous task loader communicating with the data server would be very beneficial in regard to architectural performance. 
However, the biggest bottlenecks lie at the model reads and uploads of both gradients and weights. 
Future work would be best served investigating efficient mechanisms to asynchronously have workers push gradients, primary update weights, primary push weights, and workers read weights on a layer-by-layer basis. 
This could drastically reduce the communication and waiting overhead currently implemented. 
Eliminating the bottleneck of a centralized primary node may force considerations towards ring-AllReduce approaches for gradient sharing. 

Other approaches for efficiency increase could lie at the gradient accumulation stage. 
Determining the effects and overcoming these issues at larger accumulation rounds could increase the performance of epoch completion. 
However, it is unknown how these larger accumulation rounds would affect convergence given the current experiments. 

Naturally, implementing a training cycle using the entire training dataset would be case for future work with both the SN and PSv architectures to determine their performance over unrestricted time and cost. 
Since the SN model is able to achieve 92\% class accuracy, it would be beneficial to determine the minimum sized dataset to achieve the highest class accuracy score. The minimum sized dataset would achieve time, and computation and financial expenses to train a successful model. 
This information could give insight into similar satellite optical imagery datasets focused on object detection. 
\section{Conclusion}
    
Ship detection using optical satellite imagery is a difficult process. 
The examples of clouds, landmasses, man-made objects, and highly reflective objects pose as issues and are a cause of false positives. 
The application in this work of using a custom U-Net model that is deeper and implements bottlenecks to extract more features demonstrates an improvement over related work. 
The custom U-Net exceeds class accuracy of related works by achieving 91.87\% accuracy. 
Training this model with two different systematic approaches provided acceptable and exceeding class accuracies compared to related works. 
The SN architecture provided an excellent class accurate model exceeding that of related works. 
The PSv architectural approach, though slower than the SN architecture, provided a model that was acceptable within the training/validation dataset and exceeded the SN model approaches in the target dataset. 

%
%
%
%

\bibliographystyle{splncs04}
\bibliography{main}
\end{document}